\documentclass[11pt]{article}
\usepackage[margin=1in]{geometry}
\usepackage{amsmath, amssymb}

\title{Gated MLPs as Symmetry-Broken Rank-1 Bilinear Attention}
\author{Nathan Breslow\thanks{LLMs were used in the ideation, writing, and drafting of this paper. The author assumes full responsibility for all content.}}
\date{}

\begin{document}
\maketitle

\begin{abstract}
We show that the conventional gated MLP can be viewed as a rank-1 approximation to a bilinear attention mechanism with two distinct factors corresponding to the query and the key. We further show that moving the nonlinearity onto one factor breaks the exchange symmetry between the two factors and, for non-homogeneous activations, the inverse-scaling symmetry as well. This perspective may help explain why gated MLPs are effective in practice and inform the design of future architectures.\end{abstract}

\section{Background: Feed-Forward Layers as Key-Value Memories}

We start with the notion of an $\text{MLP}$ layer, which is applied to each token $x_i \in \mathbb{R}^d$ independently. Following the row-vector convention used by Geva et al.~\cite{geva2021transformer}, we treat $x_i$ as a row vector. The traditional two-layer FFN, dropping bias terms, can be expressed as:

$$\text{MLP}(x_i) = \sigma(x_i W_1^T) W_2$$

where $W_1 \in \mathbb{R}^{d_{ff} \times d}$ and $W_2 \in \mathbb{R}^{d_{ff} \times d}$ are the weight matrices, and $\sigma$ is a non-linear activation function applied elementwise.

Using the key-value memory interpretation of transformer feed-forward layers introduced by Geva et al.~\cite{geva2021transformer}, we can view $W_1$ as a set of $d_{ff}$ row vectors $k_j \in \mathbb{R}^d$ for $j=1, \ldots, d_{ff}$, and $W_2$ as a set of $d_{ff}$ row vectors $v_j \in \mathbb{R}^d$. Thus, we can rewrite the MLP as:

$$\text{MLP}(x_i) = \sum_{j=1}^{d_{ff}} \sigma(x_i k_j^T) v_j$$

or in matrix form:

$$\text{MLP}(x_i) = \sigma(x_i K^T) V$$

There is an additional form, introduced by Shazeer~\cite{shazeer2020glu}, that gates the output of $W_1$ with a separate linear transformation of the input:

$$\text{GatedMLP}(x_i) = \left(\sigma(x_i W_1^T) \odot (x_i W_3^T)\right) W_2$$

where $W_3 \in \mathbb{R}^{d_{ff} \times d}$ is an additional weight matrix. We will show, in this paper, a mathematical connection between the original MLP formulation and the gated MLP formulation through the intermediary of a bilinear attention-esque mechanism.

\section{From Static Keys to Bilinear Scores}

Now, one may notice the key-value structure of the conventional MLP and wish to introduce a dynamically computed query vector $q_i$ that depends on the input token $x_i$.

One may first notice that simply having a single query matrix $Q \in \mathbb{R}^{d \times d}$ that transforms $x_i$ into a query vector $q_i = x_i Q$ does not change the expressivity of the model, as:

$$\text{MLP}(x_i) = \sigma(q_i K^T) V = \sigma(x_i Q K^T) V$$

Setting $W_1^\top = Q K^\top$ (equivalently $W_1 = K Q^\top$) and $W_2 = V$ recovers the original MLP formulation.

Therefore, one may consider a separate query matrix $Q_j$ for each key-value pair, leading to:

$$\text{MLP}(x_i) = \sum_{j=1}^{d_{ff}} \sigma(x_i Q_j k_j^T) v_j$$

This however is equivalent to setting $W_1^T$ to have columns $Q_j k_j^T$ for $j=1, \ldots, d_{ff}$, and thus does not increase the expressivity of the model. The per-pair query matrices $Q_j$ can be absorbed into the key matrix $K$ without loss of generality.

Therefore, in order to unlock a new level of expressivity, we need to allow both the per-pair queries and keys to be dynamically computed from the input token - via $Q_j$ and $K_j$ respectively. This leads to the following formulation:

$$\text{MLP}(x_i) = \sum_{j=1}^{d_{ff}} \sigma(x_i Q_j (x_i K_j)^T) v_j$$

This becomes:

$$\text{MLP}(x_i) = \sum_{j=1}^{d_{ff}} \sigma(x_i Q_j K_j^T x_i^T) v_j$$

Now, this is prohibitive to store for large $d_{ff}$, as it requires storing a tensor of size $d_{ff} \times d \times d$ for the queries, and similarly for the keys. 

\section{Rank-1 Gated MLP Approximation}

So instead, let us consider a rank-1 approximation of $Q_j K_j^T$ as $q_j^T k_j$ where $q_j, k_j \in \mathbb{R}^d$ are row vectors. This leads to the following formulation:

$$\text{MLP}(x_i) = \sum_{j=1}^{d_{ff}} \sigma(x_i q_j^T k_j x_i^T) v_j$$

Now, define matrix $Q \in \mathbb{R}^{d_{ff} \times d}$ with rows $q_j$ and matrix $K \in \mathbb{R}^{d_{ff} \times d}$ with rows $k_j$. Using $\odot$ to denote elementwise multiplication, we can write:

$$\text{MLP}(x_i) = \sigma\left((x_i Q^T) \odot (x_i K^T)\right) V$$

This resembles the bilinear/GLU family of multiplicative feed-forward layers introduced by Dauphin et al.~\cite{dauphin2017language} and applied to LLMs by Shazeer~\cite{shazeer2020glu}.

\section{Higher-Rank Generalization}

Note that this mechanism can be generalized to arbitrary-rank approximations by allowing $Q_j K_j^T$ to be a sum of multiple rank-1 matrices as follows:

$$\text{MLP}(x_i) = \sum_{j=1}^{d_{ff}} \sigma\left(\sum_{r=1}^R x_i q_{jr}^T k_{jr} x_i^T\right) v_j$$

where $q_{jr}, k_{jr} \in \mathbb{R}^d$ are row vectors for $r=1, \ldots, R$.

Or in matrix form:

$$\text{MLP}(x_i) = \sigma\left(\sum_{r=1}^R (x_i Q_r^T) \odot (x_i K_r^T)\right) V$$

where $Q_r, K_r \in \mathbb{R}^{d_{ff} \times d}$ are the query and key matrices for the $r$-th rank-1 component. The sum over $r$ is taken componentwise before applying $\sigma$.

We leave further exploration of this generalization to future work. 

\section{Symmetry Breaking in Gated MLPs}

Now, we return to the rank-1 case:

$$\text{MLP}(x_i) = \sigma\left((x_i Q^T) \odot (x_i K^T)\right) V$$

This stands in contrast to the more conventional gated MLP form:

$$\text{GatedMLP}(x_i) = \left(\sigma(x_i Q^T) \odot (x_i K^T)\right) V$$

where the nonlinearity is applied to the query term before the elementwise multiplication.

The conventional form can be viewed as breaking two symmetries the original formulation has - scaling and swapping. 

To see why, return to the elementwise formulation:

$$\text{MLP}(x_i) = \sum_{j=1}^{d_{ff}} \sigma(x_i q_j^T k_j x_i^T) v_j$$

Here, if $q_j$ is scaled by some factor $\alpha$ and $k_j$ is scaled by $\frac{1}{\alpha}$, the output of the model does not change (regardless of the choice of $\sigma$):

$$\text{MLP}(x_i) = \sum_{j=1}^{d_{ff}} \sigma(x_i (\alpha q_j)^T \frac{1}{\alpha} k_j x_i^T) v_j = \sum_{j=1}^{d_{ff}} \sigma(\alpha \cdot \frac{1}{\alpha} x_i q_j^T k_j x_i^T) v_j = \sum_{j=1}^{d_{ff}} \sigma(x_i q_j^T k_j x_i^T) v_j$$

Similarly, if we swap $q_j$ and $k_j$, the output also does not change (regardless of the choice of $\sigma$):

$$\text{MLP}(x_i) = \sum_{j=1}^{d_{ff}} \sigma(x_i k_j^T q_j x_i^T) v_j = \sum_{j=1}^{d_{ff}} \sigma(x_i q_j^T k_j x_i^T) v_j$$

However, if we apply the nonlinearity to the query term before the elementwise multiplication, as below:

$$\text{GatedMLP}(x_i) = \sum_{j=1}^{d_{ff}} (\sigma(x_i q_j^T) (x_i k_j^T)) v_j$$

these symmetries are generally broken. Scaling $q_j$ by $\alpha$ and $k_j$ by $\frac{1}{\alpha}$ can now change the output:

$$\text{GatedMLP}(x_i) = \sum_{j=1}^{d_{ff}} (\sigma(x_i (\alpha q_j)^T) (x_i \frac{1}{\alpha} k_j^T)) v_j$$

$$= \sum_{j=1}^{d_{ff}} (\sigma(\alpha x_i q_j^T) \frac{1}{\alpha} (x_i k_j^T)) v_j$$

If $\sigma$ is non-homogeneous, $\sigma(\alpha x_i q_j^T)$ is not equal to $\alpha \sigma(x_i q_j^T)$ in general, so the output changes. Note that ReLU is positively homogeneous, so this does not hold for positive scaling factors $\alpha$; for other activations like SiLU or GELU, the inverse-scaling symmetry is generally broken.

But even for ReLU, swapping $q_j$ and $k_j$ still generally changes the output:

$$\text{GatedMLP}(x_i) = \sum_{j=1}^{d_{ff}} (\sigma(x_i k_j^T) \odot (x_i q_j^T)) v_j$$

$$\neq \sum_{j=1}^{d_{ff}} (\sigma(x_i q_j^T) \odot (x_i k_j^T)) v_j$$

In fact, the gated form generally breaks the exchange symmetry regardless of whether $\sigma$ is homogeneous. Writing
$$
a = x_i q_j^\top, \qquad b = x_i k_j^\top,
$$
exchange symmetry would require
$$
\sigma(a)b = \sigma(b)a
$$
for all possible $a,b$. For nonzero $a,b$, this is equivalent to requiring
$$
\frac{\sigma(a)}{a} = \frac{\sigma(b)}{b},
$$
so $\sigma(z)/z$ would have to be constant. Up to degenerate cases, this forces $\sigma$ to be linear. Therefore, nonlinear activations, including ReLU, GELU, and SiLU, break the exchange symmetry in general.

So we see that the conventional gated MLP can be viewed as a variant of the rank-1 bilinear approximation, breaking the exchange symmetry between the query and key factors, and for non-homogeneous activations, the inverse-scaling symmetry as well.

\section{Conclusion}

We have shown that if one dynamically computes both the queries and keys from the input token, one can express the feed-forward layer as a bilinear attention mechanism. We further showed that the conventional gated MLP can be viewed as a rank-1 approximation to this bilinear attention, and that moving the nonlinearity onto one branch breaks the exchange symmetry between the two factors and, for non-homogeneous activations, the inverse-scaling symmetry as well. This perspective may help explain why gated MLPs are effective in practice and inform the design of future architectures.

We emphasize that this is, at present, an interpretive framework rather than an empirical one: it offers a lens for understanding the structure of gated MLPs and we make no predictions to be validated experimentally. We leave such investigation, along with the higher-rank generalization of Section 4, to future work.

\bibliographystyle{plain}
\bibliography{refs}

\end{document}